\documentclass{article}

\usepackage{microtype}
\usepackage{graphicx}
\usepackage{subcaption}
\usepackage{booktabs}
\usepackage{hyperref}
\usepackage{xspace}

\newcommand{\eg}{\textit{e.g.,}\xspace}

\usepackage[preprint]{icml2026}

\usepackage{amsmath}
\usepackage{amssymb}
\usepackage{mathtools}
\usepackage{amsthm}

\usepackage[capitalize,noabbrev]{cleveref}

\theoremstyle{plain}

\theoremstyle{definition}

\theoremstyle{remark}

\usepackage[textsize=tiny]{todonotes}

\usepackage{ifthen}
\usepackage{xcolor}

\usepackage{svg}

\usepackage{tcolorbox}
\tcbuselibrary{breakable,skins}

\newtcolorbox{promptbox}[2][]{
  enhanced,
  breakable,
  colback=#2!5!white,
  colframe=#2!75!black,
  fonttitle=\bfseries,
  title=#1,
  left=6pt,
  right=6pt,
  top=6pt,
  bottom=6pt,
  boxrule=0.5pt,
  arc=2pt,
}

\definecolor{codingagentcolor}{RGB}{52, 119, 181}
\definecolor{reactcolor}{RGB}{181, 92, 52}
\definecolor{fullcontextcolor}{RGB}{92, 140, 92}
\definecolor{retrievercolor}{RGB}{140, 92, 140}

\icmltitlerunning{Coding Agents are Effective Long-Context Processors}

\begin{document}

\twocolumn[
  \icmltitle{Coding Agents are Effective Long-Context Processors}

  \icmlsetsymbol{equal}{*}

  \begin{icmlauthorlist}
    \icmlauthor{Weili Cao}{duke}
    \icmlauthor{Xunjian Yin}{duke}
    \icmlauthor{Bhuwan Dhingra}{equal,duke}
    \icmlauthor{Shuyan Zhou}{equal,duke}
  \end{icmlauthorlist}

  \icmlaffiliation{duke}{Duke University, Durham, NC, USA}
  
  \icmlcorrespondingauthor{Weili Cao}{weili.cao@duke.edu}

  \icmlkeywords{LLM Agent, Long Context, Coding Agent}

  \vskip 0.3in
]

\printAffiliationsAndNotice{\icmlEqualContribution}

\begin{abstract}
Large Language Models (LLMs) have demonstrated remarkable progress in scaling to \emph{access} massive contexts.
However, the access is via the latent and uninterpretable attention mechanisms, and LLMs fail to effective \emph{process} long context, exhibiting significant performance degradation as context length increases.
In this work, we study whether long-context processing can be externalized from latent attention into explicit, executable interactions, by allowing coding agents to organize text in file systems and manipulate it using its native tools.
We evaluate off-the-shelf frontier coding agents as the general interface for tasks that require processing long contexts, including long-context reasoning, retrieval-augmented generation, and open-domain question answering with large-scale corpus contains up to \emph{three trillion} tokens.
Across multiple benchmarks, these agents outperform published state-of-the-art by 17.3\% on average.
We attribute this efficacy to two key factors: \emph{native tool proficiency}, which enables agents to leverage executable code and terminal commands rather than passive semantic queries, and \emph{file system familiarity}, which allows them to navigate massive text corpora as directory structures. 
These findings suggest that delegating long-context processing to coding agents offers an effective alternative to semantic search or context window scaling, opening new directions for long-context processing in LLMs. Our code is available at \href{https://github.com/weilicao/Coding_Agents_are_Effective_Long_Context_Processors}{this repository.}
\end{abstract}
\section{Introduction}

\begin{figure}[t]
    \centering
    \includegraphics[width=\columnwidth]{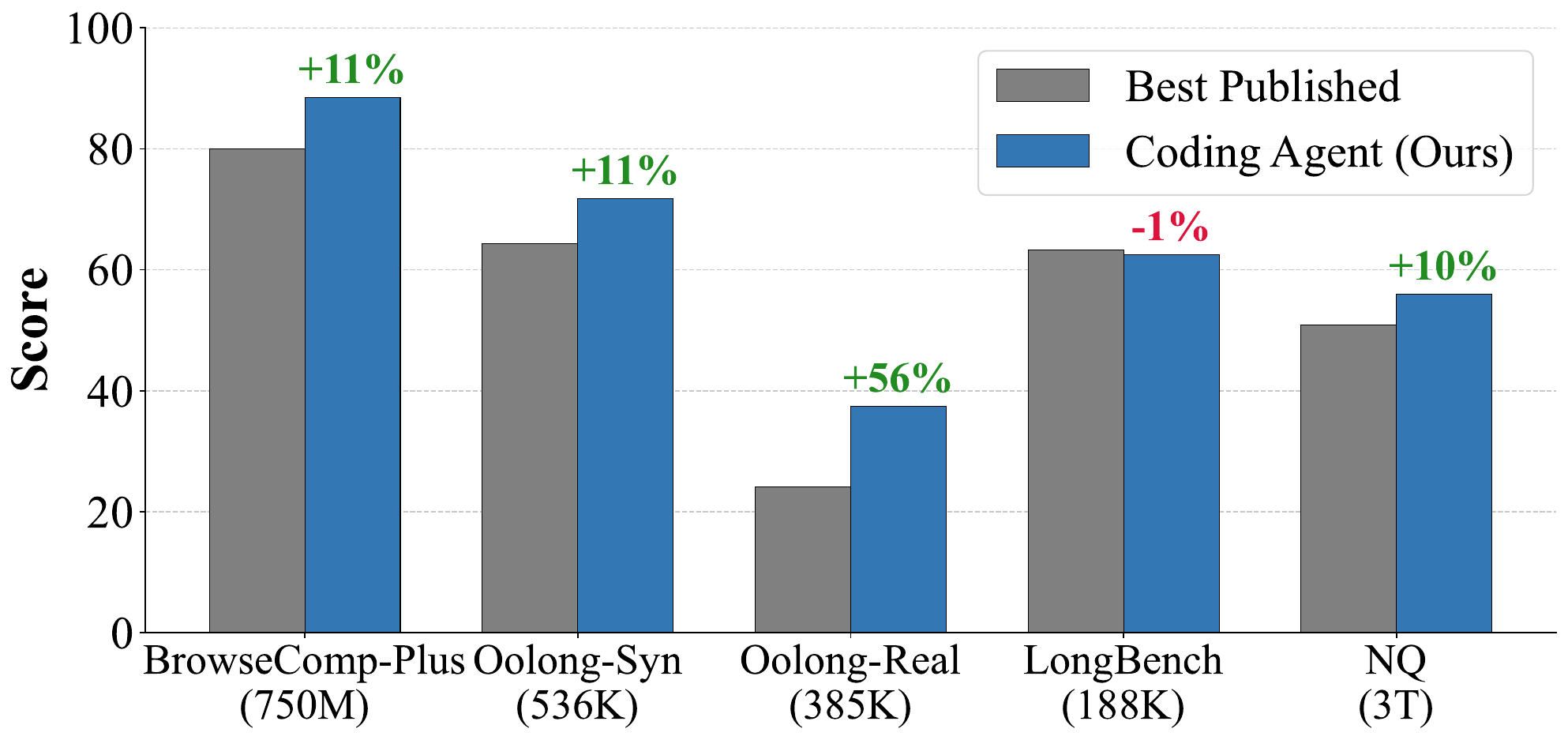}
    \caption{Coding agents significantly outperform best published results across five long-context benchmarks spanning from 188K to three trillion tokens. Green percentages indicate relative improvement over prior state-of-the-art.}
    \label{fig:main_results}
\end{figure}

\begin{figure*}[t]
    \centering
    \includegraphics[width=\columnwidth*3/2]{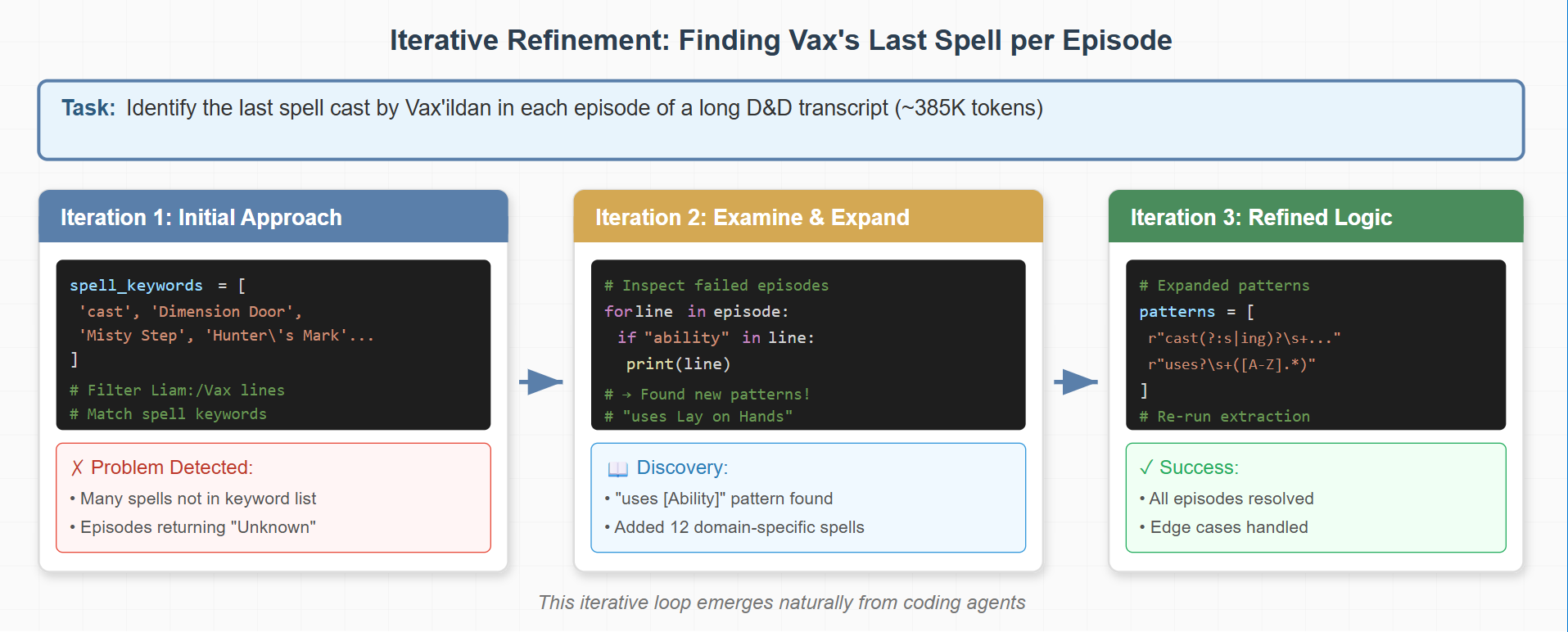}
    \caption{Iterative refinement example on Oolong-Real. When asked to identify the last spell cast by Vax'ildan in each episode of a 385K-token transcript, the coding agent wrote a Python script, discovered domain-specific spell references through failure analysis, and iteratively refined its logic.}
    \label{fig:iterative_example}
\end{figure*}

Modern applications increasingly require models to reason over massive corpora, such as scientific archives, or web-scale text collections. 
Recent advances in LLMs have significantly scaled supported context windows, with frontier systems now handling millions of tokens \cite{gemini, claude}.
Recent results show that drop-in long-context models can outperform retrieval-augmented generation (RAG) systems \cite{sps, rag_vs_lc_1, rag_vs_lc_2, lc_strong_1}, despite the extensive optimization of modern retrieval pipelines.

Despite these gains, long-context scaling primarily improves input access rather than effective processing.
As context length grows, models suffer from context rot \cite{context_rot}, with performance degrading as context length increases \cite{oolong, scitrek, LCLM_agent, loogle_v2}.
Moreover, reasoning remains latent and uninterpretable, as models provide little transparency into which parts of the context inform a given generation.
While recent work has advanced interpretability of model internals \cite{interpretability_2, interpretability_3, interpretability_5}, these methods remain difficult to apply at scale~\cite{interpretability_4}.

RAG addresses some of these challenges by externalizing long-context access through retrieval and reasoning stages.
However, standard RAG pipelines rely on fixed, shallow retrieval mechanisms, which limit their ability to support iterative, multi-hop reasoning where intermediate findings must guide subsequent queries \cite{rag_limit_1, rag_limit_2}.
As a result, RAG systems offer limited flexibility for complex long-context processing tasks such as multi-hop question answering.

In this work, we propose a different approach based on a simple observation: 
coding agents, trained on large code repositories with long files and hierarchical structure, can transfer these skills to long-context text processing tasks.
Rather than relying on latent attention or fixed retrieval, such agents can explicitly organize, filter, and transform text using executable programs.

As illustrated in \autoref{fig:iterative_example}, when asked to identify the last spell cast by a specific character in each episode of a long transcript, a coding agent wrote a script that segmented the document by episode, filtered relevant mentions, and extracted spell names using pattern matching.
When the initial script failed to capture many cases, the agent inspected intermediate outputs, discovered domain-specific spell references, and iteratively refined its logic.
This iterative, programmatic interaction is difficult to realize with fixed retrieval pipelines or passive long-context attention, but arises naturally from the agent's software engineering training.

Building on this intuition, we frame long-context processing as a file system navigation and manipulation problem with coding agents.
We place massive text corpora into directory structures and delegate processing to off-the-shelf coding agents \cite{codex_website}, which can explore and manipulate these structures using familiar tools such as terminal commands, programmatic search, file manipulation, and iterative execution.

We evaluate coding agents on long-context QA benchmarks spanning two settings:
BrowseComp-Plus \cite{browsecomp_plus} and Natural Questions \cite{nq}, which require synthesizing answers from information distributed across massive corpora; and
LongBench \cite{longbench_v2} and the Oolong benchmarks \cite{oolong}, which require reasoning over individual long documents.

As shown in \autoref{fig:main_results}, coding agents consistently outperform strong baselines across all settings, establishing new state-of-the-art results on \emph{four out of five} benchmarks and remaining competitive on the remaining one.
These gains persist across context scales ranging from hundreds of thousands to trillions of tokens, and the gain holds across different LLM backbones.

Our analysis attributes this effectiveness to two core capabilities: \emph{native tool proficiency}, which enables precise, executable interactions beyond natural-language queries, and \emph{file system familiarity}, which provides strong inductive priors for navigating large text collections.
These capabilities also help explain a surprising negative result: equipping coding agents with standard retrieval tools does not consistently improve performance.
More interestingly, we observe \emph{emergent, task-specific processing strategies}: agents autonomously develop iterative query refinement for multi-hop retrieval, programmatic aggregation for analytical tasks, and hybrid strategies for reading comprehension, all arising without explicit instruction or specialized training.

We hope the strong performance demonstrated in our work encourages a rethinking of simple, versatile approaches as backbone LLMs grow increasingly capable.

\begin{figure*}[ht!]
    \centering
    \includegraphics[width=\columnwidth*3/2]{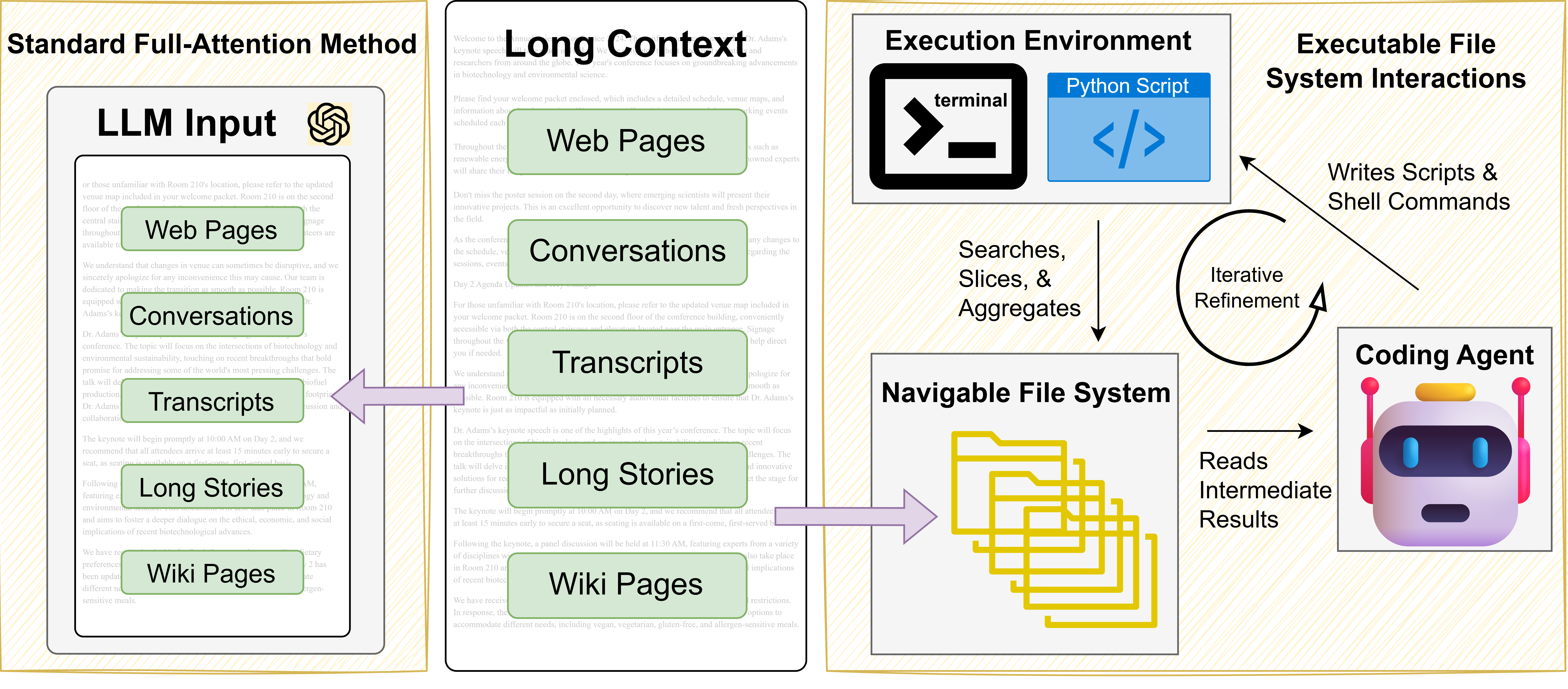}
\caption{\textbf{Text Processing as File System Navigation.} We organize text corpus into a \textbf{Navigable File System} of documents and folders. The \textbf{Coding Agent} explores this hierarchy using native tools (e.g., \texttt{ripgrep}, terminal commands), writes Python scripts for \textbf{Programmatic Aggregation}, and saves intermediate results. The agent \textbf{Iteratively Refines} its queries based on discovered information, enabling multi-hop reasoning without fixed retrieval pipelines.}
\label{fig:main_figure}
\end{figure*}

\section{Text Processing as File System Operation}

Our approach reformulates long-context processing as a file system operation task, as illustrated in \autoref{fig:main_figure}. Rather than feeding massive text directly into a model's context window or relying on semantic retrieval, we structure textual content as files within a directory hierarchy and delegate processing to off-the-shelf coding agents. 

\textbf{Problem Setup.} Given a query $q$ and either a large corpus $\mathcal{C} = \{d_1, d_2, \ldots, d_n\}$ where $d_i$ is a piece of document, or a single long document $D$, the task is to produce an answer $a$ by reasoning over the provided corpus.

\textbf{Corpus Formatting.} For large corpus settings (corpus size $>$ 100M tokens), we format each document as an individual \texttt{txt} file and organize these files within a corpus directory.\footnote{For NQ, since the corpus is prohibitively large, we store all documents in a single JSONL file.} For single long-document settings in long-context QA tasks, we place the entire context in one txt file.

\textbf{Agent Interface.} The coding agent receives only the file or directory path along with the query. The agent then freely employs its native capabilities: executing terminal commands (\eg \texttt{grep} and \texttt{head}), writing and running Python scripts for programmatic search and text processing, creating intermediate files to store partial results, and iteratively refining its exploration based on discovered information.

Crucially, we impose no constraints on how the agent processes the content. The agent autonomously decides whether to scan files sequentially, construct keyword searches, write custom parsing scripts, or combine multiple strategies. In some configurations, we additionally provide agents with access to a retrieval tool (BM25 or dense embeddings); however, even in these settings, the agent retains full autonomy over whether and how to use these tools. This stands in contrast to RAG pipelines with fixed retrieval stages or ReAct agents limited to predefined tool APIs. The complete prompts for all methods are provided in \autoref{appendix:prompts}.

\begin{table*}[t]
\caption{Main results across five benchmarks. Best results are in \textbf{bold}. \textsuperscript{*} indicates results evaluated on the full set, used here only for reference. Reported metrics are Accuracy for BrowseComp-Plus and LongBench, Exact Match (EM) for NQ, and Score for Oolong.}
\label{tab:main_results}
\centering\small
\begin{tabular}{lccccc}
\toprule
\textbf{Method} & \textbf{BrowseComp-Plus} & \textbf{Oolong-Syn} & \textbf{Oolong-Real} & \textbf{LongBench} & \textbf{NQ} \\
\midrule
\textit{Context Length} & 750M & 536K & 385K & 188K & 3T \\
\midrule
GPT-5 Full Context & 20.00 & 59.22 & 22.45 & 61.00 & 27.00 \\
RAG & 65.00 & 45.53 & 13.38 & 50.50 & 47.00 \\
ReAct Agent & 72.50 & 31.39 & 19.06 & 59.00 & 49.00 \\
RLM & -- & 64.38 & 23.07 & 54.00 & 55.33 \\
Best Published & 80.00\rlap{\textsuperscript{*a}} & 64.38\rlap{\textsuperscript{b}} & 24.09\rlap{\textsuperscript{c}} & \textbf{63.30}\rlap{\textsuperscript{*d}} & 50.90\rlap{\textsuperscript{*e}} \\
\midrule
\multicolumn{6}{c}{\textit{Coding Agents} (Ours)} \\
\midrule
Codex (No Retriever) & \textbf{88.50} & \textbf{71.75} & 33.73 & 61.50 & \textbf{56.00} \\
Codex + Gemini Emb. & 84.00 & 68.03 & 32.40 & 61.50 & -- \\
Codex + BM25 & 78.50 & 71.07 & 30.86 & 60.80 & 53.00 \\
Claude Code + BM25 & -- & -- & \textbf{37.46} & \textbf{62.50} & -- \\
\bottomrule
\end{tabular}\\[2pt]
{\footnotesize \textsuperscript{a}\cite{openJiuwen}, \textsuperscript{b}\cite{rlm}, \textsuperscript{c}\cite{gpt5}, \textsuperscript{d}\cite{gemini}, \textsuperscript{e}\cite{interact_rag}}
\end{table*}

\section{Experiments}

\subsection{Benchmarks}

\textbf{BrowseComp-Plus} \cite{browsecomp_plus} is a web browsing benchmark for evaluating Deep-Research agents on complex, multi-hop question answering. Built upon BrowseComp \cite{browsecomp}, BrowseComp-Plus provides a fixed corpus of 100K web documents and guarantees to contain the gold documents. The task requires agents to iteratively search and reason across multiple documents to locate hard-to-find, entangled information. For the evaluation, we employ an LLM-as-a-judge approach using GPT-5 to assess whether predicted answers match the ground truth, and report accuracy.

\textbf{LongBench-v2} \cite{longbench_v2} is a long-context benchmark designed to evaluate the ability of LLMs to perform deep understanding and complex reasoning across diverse real-world tasks. The benchmark adopts a multiple-choice question answering format and encompasses six task categories: single-document QA, multi-document QA, long in-context learning, long-dialogue history understanding, code repository understanding, and long structured data understanding. We report accuracy for these MCQs.

\textbf{Oolong-Real} and \textbf{Oolong-Synthetic} are two variants from the Oolong \cite{oolong} benchmark designed for long-context reasoning. Oolong tasks require analyzing, synthesizing, and aggregating information distributed across entire documents to answer questions about patterns and distributions. Both variants test models' ability to reason over large quantities of examples, perform in-context classification and counting, and handle temporal and user relations. we follow the scoring protocol described in the original paper for evaluation: questions requiring a label, date, user ID, or comparison are scored using exact match, and questions requiring a numerical answer are scored using $\texttt{score}(\hat{y}) = 0.75^{|y - \hat{y}|}$. 

\textbf{Natural Questions (NQ)} \cite{nq} is a widely-used open-domain question answering benchmark. The task requires retrieving relevant passages from a large-scale Wikipedia corpus and extracting short factoid answers. We report exact match (EM) after normalization.

Due to computational cost, we randomly sample 200 examples from each benchmark, and we rerun all baselines with the same subset for fair comparisons.

\subsection{Baselines}

\textbf{LLM full-context} We evaluate the ability of GPT-5 \cite{gpt5} to directly answer questions given full context (see \autoref{appendix:fullcontext_prompts} for prompt templates). For BrowseComp-Plus and NQ, since the corpus is too large for the LLM to handle, we randomly sample documents from the corpus to form a 100k-token context. For LongBench and Oolong, we apply a sliding window strategy following prior work \cite{sps}.
For datapoints with context lengths greater than 200k tokens, we use a window size of 200k tokens with 50k overlaps. Answers and reasoning are produced from each window and then aggregated by the same LLM, which produces a final answer.

\textbf{RAG} We follow a standard RAG pipeline: retrieve the top 10 documents (for corpus-level tasks) or 300-word chunks (for long-document tasks), then generate the answer using GPT-5. We use Gemini embeddings \cite{gemini_embedding} for retrieval. We use BM25 for NQ due to its large corpus size.

\textbf{ReAct-Style Search Agents} Following \cite{browsecomp_plus, context_folding}, we perform agentic search by placing the LLM in a ReAct loop \cite{react}. We provide GPT-5 with a Gemini embedding model \cite{gemini_embedding} as a retrieval tool. The LLM is shown the question and given access to ``retrieve'' and ``get document'' tools (see \autoref{appendix:react_prompts}).

\textbf{Recursive Language Model (RLM)} Recursive Language Models \cite{rlm} treat long input text as part of an external environment where LLMs can programmatically examine and recursively call themselves over text snippets using a Python REPL. We evaluate RLM using the exact setting described in the original paper. We exclude RLM from our BrowseComp-Plus evaluation because running it on the full 100k-document corpus is prohibitively time-consuming; notably, the original RLM paper evaluates only on a reduced 1,000-document subset of this benchmark.

\subsection{Coding Agent}

We evaluate Codex v0.46.0 \cite{codex_website}
with GPT-5 as the base model under three configurations: (1) Native codex without any retriever, (2) Codex with BM25 as the retriever, and (3) Codex with dense retriever using Gemini embeddings as the encoder. We use the default system prompt in the first setting.
We include instructions explaining how to use the retriever, along with the retriever's Python implementation in the second and the third setting (see Appendix \ref{appendix:coding_agent_prompts} for the complete prompts).
For retrieval of Longbench and Oolong, we split documents into chunks of 300 words following prior work \cite{rag_vs_lc_1, rag_vs_lc_2}.

We additionally evaluate Claude Code with Sonnet 4.5 \cite{claude} as the base model. The purpose of this experiment is to demonstrate that our findings are not specific to a single coding agent implementation. Claude Code represents an alternative frontier coding agent with distinct training and architecture from Codex.\footnote{Due to budget constraints, we limit our Claude Code evaluation to two benchmarks: Oolong-Real and LongBench.}

\begin{table}[t]
\caption{Ablation study on file system structure. We test Codex with GPT-5 on BrowseComp-Plus.}
\label{tab:file_system_ablation}
\centering\small
\begin{tabular}{lcc}
\toprule
\textbf{Retriever} & \textbf{Folder Structure} & \textbf{Single File} \\
\midrule
No Retriever & 89.0 & 83.0 \\
Gemini Emb. & 90.0 & 86.0 \\
BM25 & 82.0 & 82.0 \\
\bottomrule
\end{tabular}
\end{table}

\section{Main Results}

\textbf{Coding Agents Establish New State-of-the-Art} As shown in \autoref{tab:main_results}, off-the-shelf coding agents significantly outperform all baselines across diverse benchmarks. Notably, these gains hold across vastly different context scales, from average 188K tokens (LongBench) to over three trillion tokens (NQ), demonstrating that coding agents provide a robust, general-purpose solution for long-context processing without task-specific training or architectural modifications.

Although GPT-5 full-context sees only a very small fraction of the corpus in its context window on large-corpus tasks, its non-trivial accuracy (20.0\% on BrowseComp-Plus, 27.0\% on NQ) is likely due to data contamination \citep{reseek}.
On Oolong, GPT-5's scores are substantially lower than reported in the original paper, which evaluated only on datapoints with context lengths under 200K tokens; while our sample includes much longer contexts where performance degrades significantly. We also note that the original RLM paper evaluates on the \texttt{trec\_coarse} subset of Oolong-Synthetic rather than the full dataset.

\begin{figure*}[t]
\centering
\small
\begin{minipage}[t]{0.496\textwidth}
\begin{tcolorbox}[enhanced,colback=codingagentcolor!5!white,colframe=codingagentcolor!75!black,fonttitle=\bfseries\scriptsize,title={\scriptsize Example 1: Iterative Query Refinement},left=3pt,right=3pt,top=2pt,bottom=2pt,boxrule=0.4pt,arc=1pt]
\scriptsize\ttfamily
\textbf{Task:} Find a pro gamer: (1) born 2000--2010, (2) game released 2020, (3) developer founded 2000--2010, (4) founder's spouse: parenting expert.\\[1pt]
\textnormal{\textit{Trace:}}\\[1pt]
\textcolor{gray}{1.} \textnormal{\textbf{Initial:}} rg "Riot Games|Brandon Beck"\\
\hspace*{2mm}$\rightarrow$ \textbf{Discover}: \textit{Brandon Beck (Riot co-founder)}\\[1pt]
\textcolor{gray}{2.} \textnormal{\textbf{Refine:}} rg "Brandon Beck.*wife"\\
\hspace*{2mm}$\rightarrow$ \textbf{Discover}: \textit{Natasha Beck (spouse)}\\[1pt]
\hspace*{2mm}\textcolor{gray}{...}\\[1pt]
\textcolor{gray}{6.} \textnormal{\textbf{Refine:}} rg "Valorant.*player|pro gamer"\\
\hspace*{2mm}$\rightarrow$ \textbf{Discover}: \textit{Demon1 (Valorant pro)}\\[1pt]
\textcolor{gray}{7.} \textnormal{\textbf{Verify:}} sed file 23043.txt\\
\hspace*{2mm}$\rightarrow$ \textbf{Confirm}: \textit{Max Mazanov, born 2002}\\[1pt]
\textbf{Answer:} Max Mazanov
\end{tcolorbox}
\end{minipage}%
\hfill
\begin{minipage}[t]{0.496\textwidth}
\begin{tcolorbox}[enhanced,colback=codingagentcolor!5!white,colframe=codingagentcolor!75!black,fonttitle=\bfseries\scriptsize,title={\scriptsize Example 2: Programmatic Aggregation},left=3pt,right=3pt,top=2pt,bottom=2pt,boxrule=0.4pt,arc=1pt]
\scriptsize\ttfamily
\textbf{Task:} Which user has most ``contradiction'' labels across 1,772 sentence pairs?\\[1pt]
\textnormal{\textit{Trace:}}\\[1pt]
\textcolor{gray}{1.} \textbf{Scan}: Parse structure $\rightarrow$ Date, User, Instance\\[1pt]
\textcolor{gray}{2.} \textbf{Write regex patterns and python function:}\\
\textcolor{blue}{neg\_re} = re.compile(r"no|not|never")\\
\textcolor{blue}{def} contradict(p, h):\\
\hspace*{1mm}\textcolor{blue}{if} neg\_re.search(h): \textcolor{blue}{return} True\\[1pt]
\textcolor{gray}{3.} \textbf{Execute \& examine:}\\
\hspace*{1mm}$\rightarrow$ \textnormal{Top users: 46054(8), 23934(12)}\\[1pt]
\textcolor{gray}{4.} \textbf{Refine patterns:} Add patterns for quantity mismatch\\
\hspace*{1mm}\textcolor{blue}{only\_re} = re.compile(r"only one|single")\\[1pt]
\textbf{Answer:} User 23934
\end{tcolorbox}
\end{minipage}%
\hfill

\caption{\small Two emergent processing strategies. \textbf{Left:} BrowseComp-Plus---iterative query refinement with entity chaining across searches. \textbf{Right:} Oolong-Synthetic---programmatic aggregation via Python scripts with regex patterns. See \autoref{appendix:detailed_traces} for detailed traces.}
\label{fig:emergent_strategies}
\end{figure*}

\section{Ablations and Analysis}
In this section, we conduct detailed ablation studies to identify the key factors contributing to the effectiveness of coding agents in long-context processing, along with an in-depth analysis of their emergent behaviors.
\subsection{File System Structure Matters}

We hypothesize that coding agents benefit from \textit{file system familiarity}, the ability to leverage directory structures acquired through training on code repositories. To test this hypothesis, we conduct an ablation study comparing two corpus organization strategies on a 100-example subset of BrowseComp-Plus.

For \textbf{Folder Structure:} Documents are organized as individual files within a directory hierarchy, mirroring the structure of typical code repositories.
For \textbf{Single File:} The directory structure is eliminated, and the corpus is stored as a single JSON dictionary where document ids serve as keys. This allows the retriever to directly output the relevant document id.

As shown in \autoref{tab:file_system_ablation}, the folder structure outperforms the single file configuration across retriever settings.

\begin{table}[t]
\caption{Analysis of average command usage counts on the BrowseComp-Plus dataset (No Retriever).
}
\label{tab:command_usage}
\centering
\begin{small}
\begin{tabular}{lccc}
\toprule
\textbf{Structure} & \textbf{Search} (\texttt{rg}) & \textbf{Extract} (\texttt{sed}) & \textbf{Index} (\texttt{nl}) \\
\midrule
Single File      & 31.05 & 0.61 & 0.69 \\
Folder & 18.46 & 4.48 & 1.22 \\
\midrule
\textit{Diff (\%)} & \textcolor{green!60!black}{-40.5\%} & \textcolor{blue}{+634\%} & \textcolor{blue}{+76.8\%} \\
\bottomrule
\end{tabular}
\end{small}
\end{table}

To understand the root cause, we compare command usage between the two structures in \autoref{tab:command_usage}, focusing on the no-retriever setting to eliminate the confounding factor of retriever tool.

With folder structure, agents employ coordinate-based reading, using \texttt{nl} (number lines) to index content and \texttt{sed} to extract specific line ranges. The usage of \texttt{sed} increases by over seven times in the folder setting, indicating that agents selectively read relevant context rather than consuming entire files. This "index and slice" strategy effectively builds a coordinate system (file + line number) that enables more accurate data extraction.
In contrast, without navigable structure, agents fall into repeated discovery loops. The higher usage of \texttt{rg} suggests that agents struggle to isolate information and must rely on expensive corpus-wide scans.

\subsection{Retrieval Tools Do Not Uniformly Improve Performance}
\label{sec: retriever_analysis}

Our main results in \autoref{tab:main_results} reveal a counterintuitive finding: equipping coding agents with retrieval tools does not consistently improve performance and can even degrade it. To better understand this phenomenon, we analyze agent behavior across retriever configurations on BrowseComp-Plus.
For each trajectory, we count native search commands measured by the number of shell commands invoking search utilities (\texttt{grep}, \texttt{ripgrep}, \texttt{find}, etc.) that do not involve the provided retriever.

\begin{table}[h]
\caption{Agent exploration patterns across retriever configurations on BrowseComp-Plus.}
\label{tab:retriever_behavior}
\centering\small
\begin{tabular}{lcc}
\toprule
\textbf{Configuration} & \textbf{Native Search} \\
\midrule
No Retriever & 14.92\\
BM25 & 9.84 \\
Gemini Emb. & 8.33 \\
\bottomrule
\end{tabular}
\end{table}

As shown in \autoref{tab:retriever_behavior}, agents without retrieval tools issue substantially more native search commands compared to retriever-augmented variants. This difference reveals a behavioral shift: when provided with a retriever, agents reduce their use of native exploration tools such as \texttt{grep}.

Counterintuitively, equipping agents with IR tools does not guarantee improved performance. We hypothesize that standard retrievers, when available, become the agent's default discovery mechanism and displace the broader file-system exploration strategies that agents otherwise employ autonomously. Since retrieval ranking is imperfect, this substitution can cause agents to miss relevant context. The precise mechanism remains an open question we leave to future work.

\subsection{Emergent Task-Specific Processing Strategies}
\label{sec:emergent_strategies}

\begin{figure}[t]
    \centering
    \includegraphics[width=0.45\textwidth]{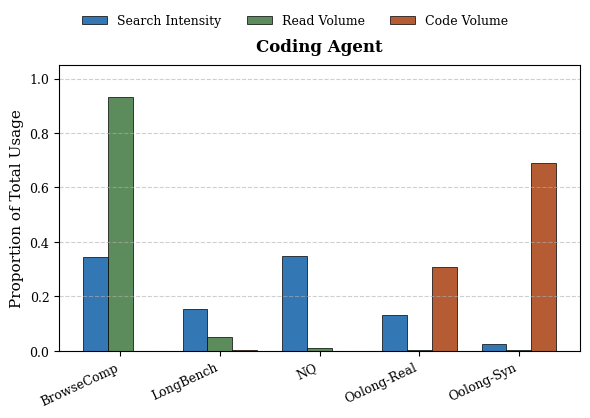}
    \caption{Quantitative characterization of agent strategies per query. The y-axis represents the normalized proportion of each metric, where the values for a given model sum to 1 across all datasets.}
    \label{fig:strategy_characterization}
\end{figure}

A key advantage of coding agents over fixed-pipeline approaches is their ability to adapt processing strategies to task requirements. We analyze agent trajectories across benchmarks and identify distinct behavioral patterns that emerge in response to different tasks.
To validate that coding agents dynamically adapt their strategies to different task types, we track Search Intensity (using search commands such as \texttt{grep} or \texttt{find}), Read Volume (number of tokens the agent reads from documents), and Code Volume (number of Python functions generated). We conduct this ablation study on codex without a retriever to eliminate confounding factors. For RLM, we count search intensity by identifying code blocks that scan files with regex patterns, excluding regex used for computation such as data processing and aggregation. The results are shown in \autoref{fig:strategy_characterization}.

\textbf{Iterative Query Refinement for Multi-Hop Retrieval.} 
On BrowseComp-Plus, which requires multi-hop reasoning across a large corpus, agents exhibit an iterative search-and-refine pattern. The agent usually begins with an initial search based on entities or concepts in the question, examines the retrieved documents, extracts new entities or relationships, and formulates refined queries targeting the next reasoning hop. Critically, this behavior emerges without explicit instruction.

\autoref{fig:emergent_strategies} (left) illustrates this pattern on a representative example. The task requires finding a professional gamer satisfying multiple constraints linked through a chain of entities. The agent begins by searching for game developers founded in the specified time range, discovering Brandon Beck as a Riot Games co-founder. It then refines its query to search for Beck's spouse, discovering Natasha Beck. Subsequent searches verify her credentials and trace the chain back to Valorant professional players, ultimately identifying Max Mazanov as the answer. This six-hop reasoning chain—Riot Games → Brandon Beck → Natasha Beck → Pepperdine → Valorant → Demon1 → Max Mazanov—emerges entirely from the agent's autonomous query refinement, with each search informed by entities discovered in previous steps.

This example reflects a broader pattern we observe quantitatively across the benchmark. As shown in \autoref{fig:strategy_characterization}, the agent prioritizes discovery over generation: BrowseComp-Plus elicits the highest Search Intensity. The agent relies primarily on an iterative loop of native search commands to locate and read relevant files.

\textbf{Programmatic Aggregation for Analytical Tasks.}
Oolong tasks require analyzing, synthesizing, and aggregating information distributed across entire documents. As shown in \autoref{fig:strategy_characterization}, on analytical tasks requiring aggregation (e.g., counting, sorting), the agent abandons search in favor of code generation. Both Oolong tasks show a dramatic drop in reading but a substantial spike in Code Volume.

\autoref{fig:emergent_strategies} (right) demonstrates this strategy on a task requiring the agent to identify which user has the most "contradiction" labels across 1,772 sentence pairs, without labels being provided. The agent writes a Python script that: (1) parses the document structure to extract user IDs and sentence pairs, (2) implements a rule-based NLI classifier using regex patterns to detect negation (no, not, never) and quantity mismatches (only one vs. series of), (3) executes the classifier over all pairs, and (4) aggregates results by user. When initial patterns miss edge cases, the agent examines intermediate outputs, expands its pattern set, and re-executes. This is an iterative refinement loop applied to code rather than queries.This approach leverages the agent's native proficiency with text processing tools. We provide concrete examples and analysis of agent-generated scripts in \autoref{appendix:case_studies}.

\textbf{Direct Inference for Diverse Long-Context Tasks.} LongBench presents a diverse mixture of long-context challenges that resist any single processing paradigm. It contains single-document and multi-document question answering, summarization, few-shot learning, synthetic retrieval, and code completion. As shown in \autoref{fig:strategy_characterization}, the coding agent has relatively low overall tool usage: modest Search Intensity, very low Read Volume, and near-zero Code Volume. This differs from the heavily read-dominated pattern on BrowseComp, the search-dominated pattern on NQ, and the code-dominated pattern on Oolong. Specifically, the near-zero Code Volume indicate that programmatic data processing is largely unnecessary for LongBench. Instead, the most effective strategy is to rely directly on the LLM's inherent long-context reasoning abilities. Consistent with this behavioral profile, our results in \autoref{tab:main_results} demonstrate that the agent's performance is nearly identical to the baseline performance of the LLM provided with the full context.

These emergent patterns demonstrate that coding agents function as generalizable long-context processors that dynamically adjust their approach based on task demands. In contrast, ReAct agents are limited to a fixed action space defined by their tool APIs, and RLMs impose a uniform recursive decomposition strategy regardless of task structure. Coding agents face no such constraints. As shown in \autoref{fig:strategy_characterization}, agents employ markedly different tools and strategies across tasks: leveraging search utilities for retrieval-heavy benchmarks, custom scripts for aggregation tasks, and hybrid approaches for reading comprehension.

\begin{table*}[t]
\caption{Average cost per query across benchmarks.}
\label{tab:cost_analysis}
\centering\small
\begin{tabular}{lccccc}
\toprule
Method & BrowseComp-Plus & Oolong-Syn & Oolong-Real & LongBench & NQ \\
\midrule
\multicolumn{6}{l}{\textit{Baselines}} \\
GPT-5 Full Context & \$0.275 & \$1.421 & \$0.770 & \$0.432 & \$0.129 \\
RAG & \$0.111 & \$0.045 & \$0.026 & \$0.024 & \$0.006 \\
ReAct Agent & \$0.237 & \$0.092 & \$0.168 & \$0.056 & \$0.027 \\
RLM & -- & \$0.920 & \$0.094 & \$0.360 & \$0.630 \\
\midrule
\multicolumn{6}{l}{\textit{Coding Agents}} \\
Codex (No Retriever) & \$0.703 & \$0.194 & \$0.419 & \$0.128 & \$0.111 \\
Codex + Gemini Emb. & \$0.628 & \$0.149 & \$0.371 & \$0.129 & -- \\
Codex + BM25 & \$0.828 & \$0.161 & \$0.368 & \$0.124 & \$0.094 \\
Claude Code + BM25 & -- & -- & \$0.380 & \$0.319 & -- \\
\bottomrule
\end{tabular}
\end{table*}

\subsection{Cost Analysis}

\autoref{tab:cost_analysis} presents the average cost per query across all benchmarks. While coding agents incur higher costs than lightweight baselines such as RAG, they remain competitive with or cheaper than other strong methods while delivering substantially superior performance.

\section{Related Work}

\textbf{Long-Context Language Models (LCLM)}
Recent advances have dramatically expanded the context windows of frontier models \cite{gpt5, claude, gemini}. This scaling has enabled direct processing of long documents. However, prior work has shown substantial performance degradation as context length increases, with models often losing much of their short-context capability well before reaching advertised limits \cite{lost_in_the_middle, context_rot, llm_lost, oolong, scitrek, LCLM_agent, loogle_v2}. \citeauthor{context_hurt} further show that context length alone can degrade performance, even with perfect retrieval quality. Additionally, inference cost of LCLMs scales linearly with context length, making very long contexts computationally expensive. These findings motivate our exploration of alternatives to context window scaling.

\textbf{Agentic RAG}
Traditional RAG methods retrieve relevant passages using a fixed pipeline, typically dense retrieval followed by answer generation, which limits their ability to handle queries requiring iterative refinement or multi-hop reasoning \cite{rankrag, chatqa}. Agentic RAG approaches address this limitation by allowing models to dynamically reformulate queries and iteratively search based on intermediate findings \cite{interact_rag, dynamicrag, search_r1, CoRAG}. However, existing agentic RAG systems are predominantly trained for specialized web search or open-domain QA tasks, requiring task-specific fine-tuning or reinforcement learning to learn effective search strategies. Our work demonstrates that off-the-shelf coding agents without any task-specific training are already capable agentic searchers.

\textbf{Agent with Long-Term Memory}
A growing body of work has focused on building and optimizing memory-centric agentic architectures through various memory manipulation strategies \cite{mem_agent_0, mem_agent_1, mem_agent_2, mem_agent_3}. \citeauthor{context_folding, agentfold} optimize memory for web search agents by folding unnecessary content. This line of work is orthogonal to ours: rather than storing context in the agent's memory, we place it in the environment as files that the agent can interact with.

Closest and concurrent to our work: Recursive Language Models (RLM) \cite{rlm} propose treating long input text as part of an external environment where LLMs can programmatically examine, decompose, and recursively call themselves over snippets of the text using a Python REPL. In principle, the two approaches share the same core intuition: rather than scaling context windows, both treat long text as an external environment that the model actively explores through a sequence of actions. The key distinction lies in how agents interact with this environment: RLMs employ a specialized system prompt that instructs models to decompose problems through recursive LLM sub-calls over text segments, whereas we use off-the-shelf coding agents with no task-specific prompting. Our agents instead leverage native file system tools (e.g., \texttt{grep}, \texttt{sed}) and custom scripts for exploration and aggregation.

\textbf{Coding Agents}
Prior work has demonstrated that incorporating coding data during fine-tuning improves LLM reasoning capabilities \cite{code_reason_1, code_reason_2, code_reason_3, code_reason_4}. \citeauthor{code_act, code_adapt} equip LLMs with code execution to solve complex reasoning tasks. However, \citeauthor{rlm} show that these agents perform poorly on long-context processing tasks. 

A another line of work trains or builds coding agents for software engineering tasks involving large codebases \cite{coding_agent_1, coding_agent_2, coding_agent_3, coding_agent_4}. These agents are designed and evaluated for long-horizon coding tasks rather than general text processing.

\section{Conclusion and Future Work}

We have demonstrated that off-the-shelf coding agents provide an effective paradigm for long-context processing, achieving state-of-the-art results on four out of five benchmarks spanning context lengths from 188K to three trillion tokens. By reformulating long-context tasks as file system navigation problems, coding agents can leverage their native capabilities, terminal commands, programmatic search, and iterative script refinement, to process massive text corpora without task-specific training or architectural modifications.

Our analysis reveals two key factors underlying this effectiveness: \emph{native tool proficiency}, which enables precise, executable interactions that go beyond natural language retrieval queries, and \emph{file system familiarity}, which provides strong inductive priors for navigating hierarchically organized text. We further observe that coding agents autonomously develop task-appropriate strategies, including iterative query refinement for multi-hop reasoning, programmatic aggregation for analytical tasks, and hybrid approaches for reading comprehension.

These findings suggest that increasingly capable foundation models for software engineering reduce the distinction between coding and general text processing tasks. Rather than relying on specialized architectures for long-context understanding, our results show that structuring text in formats aligned with code can be sufficient for effective reasoning over extended contexts.

Our approach has several limitations that suggest directions for future work. First, our analysis reveals that naively providing retrieval tools may degrade performance; future work should investigate how to better integrate retrieval capabilities without suppressing agents' native exploration. Second, while off-the-shelf coding agents transfer surprisingly well to text processing tasks, they are primarily aligned and optimized for coding rather than long-context reasoning.

An important direction for future work is developing frameworks that specialize these agents for navigating and reasoning over massive text corpora.

\section*{Impact Statement}

This paper presents work whose goal is to advance the field of Machine Learning. There are many potential societal consequences of our work, none which we feel must be specifically highlighted here.

\section*{Acknowledgments}

This work is partially supported by the Learning Engineering Virtual Institute, funded by leading education philanthropists and organizations through Grant G-23-2137070 to the University of Florida and its partner institutions. This work is also supported by Google.org, the Google Cloud Research Credits program for the Gemini Academic Program, and Amazon AGI Labs SF.

\newpage

\bibliography{example_paper}
\bibliographystyle{icml2026}

\newpage
\appendix
\onecolumn

\section{Prompts}
\label{appendix:prompts}

This section provides the complete prompts used for all methods evaluated in our experiments. We organize prompts by method type and benchmark. Variables in curly braces (e.g., \texttt{\{question\}}, \texttt{\{context\_location\}}) are replaced with actual values at runtime.

\subsection{Coding Agent Prompts}
\label{appendix:coding_agent_prompts}

We present prompts for our coding agent approach under two configurations: (1) without retriever access, where agents rely entirely on native file system exploration, and (2) with retriever access, where agents can optionally use a retrieval tool alongside their native capabilities.

\subsubsection{Without Retriever}

\begin{promptbox}[BrowseComp-Plus]{codingagentcolor}
\small\ttfamily
Answer the following question by iterating through files under folder \{context\_location\}:

Question:\\
\{question\}

BrowseCompPlus\_Corpus contains 100k files, and some of the files contain information needed to answer this question.
\end{promptbox}

\begin{promptbox}[LongBench-v2]{codingagentcolor}
\small\ttfamily
Answer the question below using text from file \{context\_location\}

Question:\\
\{Question\}

Choices:\\
(A) \{Choice\_A\}\\
(B) \{Choice\_B\}\\
(C) \{Choice\_C\}\\
(D) \{Choice\_D\}

Choose only one option.

Format your response as follows: "The correct answer is (insert answer here)".
\end{promptbox}

\begin{promptbox}[Natural Questions]{codingagentcolor}
\small\ttfamily
Answer the following question by iterating through the corpus \{context\_location\}

Question:\\
\{question\}

Answer the question based on the given document. Only give me the answer and do not output any other words.
\end{promptbox}

\begin{promptbox}[Oolong-Real]{codingagentcolor}
\small\ttfamily
Answer the following question using text from file\\
\{context\_location\}

The description of the txt file is included at the start of the file.

Question: \{question\}
\end{promptbox}

\begin{promptbox}[Oolong-Synthetic]{codingagentcolor}
\small\ttfamily
Answer the following question using text from file\\
\{context\_location\}

The description of the txt file is included at the start of the file.

Question: \{question\}

Note: Labels are not provided, you should determine label by yourself.
\end{promptbox}

\subsubsection{With Retriever}
\label{appendix:coding_agent_retriever}

When equipped with a retriever, the coding agent receives additional instructions explaining how to invoke the retrieval tool. The \texttt{\{embedding\_model\}} parameter is set to either \texttt{BM25} or \texttt{Gemini Emb.} depending on the retriever configuration.

\begin{promptbox}[BrowseComp-Plus + Retriever]{retrievercolor}
\small\ttfamily
Answer the following question by iterating through files under folder \{context\_location\}:

Question:\\
\{question\}

BrowseCompPlus\_Corpus contains 100k files, and some of the files contain information needed to answer this question.

Here is a retriever you may use to search for documents:

python3 retriever.py --dataset browsecomp-plus --embedding-model \{embedding\_model\} --top-k 5 --query "your query here"

Keep in mind that there are 100k documents in the corpus. When using the retriever for your search, carefully format your query to be multi-faceted.
\end{promptbox}

\begin{promptbox}[LongBench-v2 + Retriever]{retrievercolor}
\small\ttfamily
Answer the question using text from file \{context\_location\}

You may use the following retriever to find top-k representative chunks (each chunk has 300 words) from the txt file:

python retriever.py --dataset longbench --embedding-model \{embedding\_model\} --query "your query here" --top-k 5 --datapoint-id \{LongBench\_datapoint\_id\}

You should identify important keywords to formulate a strong query. You can form an initial query, analyze the retrieved chunks, and then iteratively refine your query by adding, removing, or changing terms based on the results to find the most relevant context.

If the retrieved chunks do not give a clear answer to the question, you must refine your query and find another set of chunks. Keep doing this until you are confident that you find the answer to the query.

What is the correct answer to this question:\\
\{Question\}

Choices:\\
(A) \{Choice\_A\}\\
(B) \{Choice\_B\}\\
(C) \{Choice\_C\}\\
(D) \{Choice\_D\}

Choose only one option.

Format your response as follows: "The correct answer is (insert answer here)".
\end{promptbox}

\begin{promptbox}[Natural Questions + Retriever]{retrievercolor}
\small\ttfamily
Answer the following question by iterating through the corpus \{context\_location\}

Question:\\
\{question\}

You may use the following retriever to search for documents:

python3 retriever.py --dataset nq --embedding-model \{embedding\_model\} --top-k 5 --query "your query here"

Answer the question based on the given document. Only give me the answer and do not output any other words.
\end{promptbox}

\begin{promptbox}[Oolong-Real + Retriever]{retrievercolor}
\small\ttfamily
Answer the following question using text from file\\
\{context\_location\}

The description of the txt file is included at the start of the file.

Question: \{question\}

You may use the following retriever to find top-k representative chunks (each chunk has 300 words) from the txt file:

python retriever.py --dataset oolong\_real --embedding-model \{embedding\_model\} --query "your query here" --top-k 5 --datapoint-id \{oolong\_datapoint\_id\}
\end{promptbox}

\begin{promptbox}[Oolong-Synthetic + Retriever]{retrievercolor}
\small\ttfamily
Answer the following question using text from file\\
\{context\_location\}

The description of the txt file is included at the start of the file.

Question: \{question\}

Note: Labels are not provided, you should determine label by yourself.

You may use the following retriever to find top-k representative chunks (each chunk has 300 words) from the txt file:

python retriever.py --dataset oolong\_synthetic --embedding-model \{embedding\_model\} --query "your query here" --top-k 5 --datapoint-id \{oolong\_datapoint\_id\}
\end{promptbox}

\subsection{ReAct-Style Search Agent Prompts}
\label{appendix:react_prompts}

The ReAct agent is provided with two tools: \texttt{retriever} for searching the corpus using semantic embeddings, and \texttt{get\_document} for retrieving the full content of a specific document. The agent performs step-by-step reasoning interleaved with tool calls.

\begin{promptbox}[BrowseComp-Plus]{reactcolor}
\small\ttfamily
Answer the given question by interacting with a retriever, using the retriever and get\_document tools provided. Please perform reasoning and use the tools step by step, in an interleaved manner. You may use the retriever and get\_document tools multiple times.

Question: \{Question\}

BrowseCompPlus\_Corpus contains 100k files, and some of the files contain information needed to answer this question.
\end{promptbox}

\begin{promptbox}[LongBench-v2]{reactcolor}
\small\ttfamily
Answer the given question by interacting with a retriever, using the retriever and get\_document tools provided. Please perform reasoning and use the tools step by step, in an interleaved manner. You may use the retriever and get\_document tools multiple times.

Question: \{Question\}

Choices:\\
(A) \{Choice\_A\}\\
(B) \{Choice\_B\}\\
(C) \{Choice\_C\}\\
(D) \{Choice\_D\}

Choose only one option.

Format your response as follows: "The correct answer is (insert answer here)".
\end{promptbox}

\begin{promptbox}[Natural Questions]{reactcolor}
\small\ttfamily
Answer the given question by interacting with a retriever, using the retriever and get\_document tools provided. Please perform reasoning and use the tools step by step, in an interleaved manner. You may use the retriever and get\_document tools multiple times.

Question: \{Question\}

Only give me the answer and do not output any other words.
\end{promptbox}

\begin{promptbox}[Oolong-Real]{reactcolor}
\small\ttfamily
Answer the given question by interacting with a retriever, using the retriever tools provided. Please perform reasoning and use the tools step by step, in an interleaved manner. You may use the retriever and get\_document tools multiple times.

Question: \{Question\}

Do not try to guess, estimate or approximate the result. Do not ask the user for clarification or follow-up. Do step-by-step reasoning if needed. Return the final answer in \textbackslash boxed\{\}.
\end{promptbox}

\begin{promptbox}[Oolong-Synthetic]{reactcolor}
\small\ttfamily
Answer the given question by interacting with a retriever, using the retriever tools provided. Please perform reasoning and use the tools step by step, in an interleaved manner. You may use the retriever and get\_document tools multiple times.

Question: \{Question\}

Do not try to guess, estimate, or approximate the result. Calculate the exact answer given these datapoints.
\end{promptbox}

\subsection{Full-Context LLM Prompts}
\label{appendix:fullcontext_prompts}

For the full-context baseline, we provide the entire context (or a sampled/windowed portion for very large corpora) directly in the prompt. The model must answer based solely on the provided context without any tool access.

\begin{promptbox}[BrowseComp-Plus]{fullcontextcolor}
\small\ttfamily
Answer the following question based on the provided context.

Question: \{Question\}

Context: \{Context\}
\end{promptbox}

\begin{promptbox}[LongBench-v2]{fullcontextcolor}
\small\ttfamily
Answer the following question based on the provided context.

Question: \{Question\}

Choices:\\
(A) \{Choice\_A\}\\
(B) \{Choice\_B\}\\
(C) \{Choice\_C\}\\
(D) \{Choice\_D\}

Context: \{Context\}

Choose only one option.

Format your answer as follows: "The correct answer is (insert answer here)".
\end{promptbox}

\begin{promptbox}[Natural Questions]{fullcontextcolor}
\small\ttfamily
Answer the following question based on the provided context.

Question: \{Question\}

Context: \{Context\}

Only give me the answer and do not output any other words.
\end{promptbox}

\begin{promptbox}[Oolong-Real]{fullcontextcolor}
\small\ttfamily
Answer the following question based on the provided context.

Question: \{Question\}

Context: \{Context\}
\end{promptbox}

\begin{promptbox}[Oolong-Synthetic]{fullcontextcolor}
\small\ttfamily
Answer the following question based on the provided context.

Question: \{Question\}

Context: \{Context\}
\end{promptbox}

\subsection{Prompt Design Rationale}
\label{appendix:prompt_rationale}

Our prompt design reflects several key principles:

\textbf{Minimal instruction for coding agents.} We deliberately keep coding agent prompts simple, providing only the task description and file location. This allows agents to leverage their native capabilities for file system navigation and text processing without constraining their approach. The contrast with retriever-augmented prompts (which include explicit tool instructions) enables us to study how tool availability affects agent behavior.

\textbf{Task-specific output formatting.} Each prompt includes output format instructions appropriate to the benchmark's evaluation protocol. LongBench uses multiple-choice format, Oolong requires exact numerical or categorical answers, and open-domain QA benchmarks expect short factoid responses.

\textbf{Consistent structure across methods.} While the available tools differ across methods (file system access for coding agents, retrieval tools for ReAct agents, none for full-context), we maintain consistent task descriptions to enable fair comparison of the underlying approaches rather than prompt engineering differences.

\section{Case Studies: Agent-Generated Scripts}
\label{appendix:case_studies}

We present example Python scripts autonomously written by Claude Code when solving Oolong benchmark tasks. These examples illustrate the \emph{programmatic aggregation} strategy discussed in \autoref{sec:emergent_strategies}, where agents write custom code to analyze, count, and aggregate information distributed across long documents.

\subsection{Example 1: Counting Dice Rolls}
\label{appendix:case_count_rolls}

\textbf{Task:} Given a transcript of a tabletop role-playing game (Critical Role), count the number of dice rolls with a specific value and compute the percentage.

\textbf{Analysis:} The agent identifies that this task requires aggregating information scattered throughout a long transcript. Rather than attempting retrieval (which would miss many instances), the agent writes a Python script that: (1) locates episode boundaries using marker tags, (2) identifies player dialogue lines by speaker prefixes, (3) applies multiple regex patterns to capture various roll announcement formats (e.g., ``rolled a 15'', ``Natural 20'', or standalone numbers), and (4) computes statistics over all extracted values.

\begin{promptbox}[Agent-Generated Script: Roll Counting]{codingagentcolor}
\small\ttfamily
\#!/usr/bin/env python3\\
import re\\
from collections import Counter\\
\\
with open('transcript.txt', 'r') as f:\\
\hspace*{1em}all\_lines = f.readlines()\\
\\
\# Find episode boundaries\\
start\_idx, end\_idx = None, None\\
for i, line in enumerate(all\_lines):\\
\hspace*{1em}if line.strip() == '[START OF EPISODE]':\\
\hspace*{2em}start\_idx = i\\
\hspace*{1em}elif line.strip() == '[END OF EPISODE]':\\
\hspace*{2em}end\_idx = i\\
\\
lines = all\_lines[start\_idx+1:end\_idx]\\
players = ['Laura', 'Travis', 'Marisha', ...]\\
rolls = []\\
\\
for line in lines:\\
\hspace*{1em}for player in players:\\
\hspace*{2em}if line.startswith(player + ':'):\\
\hspace*{3em}dialogue = line[len(player)+1:].strip()\\
\hspace*{3em}\# Pattern: "rolled X" or "rolled a X"\\
\hspace*{3em}matches = re.findall(\\
\hspace*{4em}r'rolled\textbackslash s+(?:a\textbackslash s+)?(\textbackslash w+)',\\
\hspace*{4em}dialogue, re.IGNORECASE)\\
\hspace*{3em}for m in matches:\\
\hspace*{4em}if m.isdigit() and 1 <= int(m) <= 40:\\
\hspace*{5em}rolls.append(int(m))\\
\\
\# Compute statistics\\
total = len(rolls)\\
target\_count = sum(1 for r in rolls if r == 13)\\
print(f"Percentage: \{100*target\_count/total:.2f\}\%")
\end{promptbox}

\subsection{Example 2: Tracking Character Actions Across Episodes}
\label{appendix:case_spells}

\textbf{Task:} Identify the last spell cast by a specific character (Vax'ildan) in each episode of a multi-episode transcript.

\textbf{Analysis:} This task requires tracking character-specific actions across multiple episodes within a single long document. The agent constructs a structured approach: (1) parse the document into separate episodes using boundary markers, (2) filter lines to those involving the target character (by speaker name or character mentions), (3) identify spell-related content using keyword matching, (4) extract spell names using regex patterns and a predefined spell list, and (5) report the last occurrence per episode.

\begin{promptbox}[Agent-Generated Script: Spell Tracking]{codingagentcolor}
\small\ttfamily
\#!/usr/bin/env python3\\
import re\\
\\
def find\_vax\_spells(filepath):\\
\hspace*{1em}with open(filepath, 'r') as f:\\
\hspace*{2em}content = f.read()\\
\\
\hspace*{1em}\# Split into episodes\\
\hspace*{1em}episodes = []\\
\hspace*{1em}current\_episode, in\_episode = [], False\\
\hspace*{1em}for line in content.split('\textbackslash n'):\\
\hspace*{2em}if line.strip() == '[START OF EPISODE]':\\
\hspace*{3em}in\_episode = True\\
\hspace*{3em}current\_episode = [line]\\
\hspace*{2em}elif line.strip() == '[END OF EPISODE]':\\
\hspace*{3em}episodes.append(current\_episode)\\
\hspace*{3em}in\_episode = False\\
\hspace*{2em}elif in\_episode:\\
\hspace*{3em}current\_episode.append(line)\\
\\
\hspace*{1em}spell\_keywords = ['cast', 'Dimension Door',\\
\hspace*{2em}'Misty Step', 'Hunter\textbackslash 's Mark', ...]\\
\hspace*{1em}results = []\\
\\
\hspace*{1em}for episode in episodes:\\
\hspace*{2em}vax\_spells = []\\
\hspace*{2em}for line in episode:\\
\hspace*{3em}if line.startswith('Liam:') or 'Vax' in line:\\
\hspace*{4em}for kw in spell\_keywords:\\
\hspace*{5em}if kw.lower() in line.lower():\\
\hspace*{6em}vax\_spells.append(line)\\
\hspace*{6em}break\\
\hspace*{2em}if vax\_spells:\\
\hspace*{3em}results.append(extract\_spell(vax\_spells[-1]))\\
\hspace*{1em}return results
\end{promptbox}

\subsection{Key Observations}

These examples demonstrate several characteristics of the coding agent's approach:

\begin{enumerate}
    \item \textbf{Structured parsing:} The agent recognizes and leverages document structure (episode markers, speaker prefixes) rather than treating the text as unstructured.

    \item \textbf{Robust pattern matching:} Multiple regex patterns handle variations in how information is expressed (e.g., ``rolled 15'' vs. ``rolled a fifteen'' vs. ``Natural 20'').

    \item \textbf{Programmatic aggregation:} Instead of retrieving a few relevant passages, the agent processes the entire document systematically to ensure complete coverage.

    \item \textbf{Domain adaptation:} The agent incorporates domain knowledge (player names, spell lists, D\&D conventions) into its parsing logic.
\end{enumerate}

These behaviors emerge without explicit instruction, demonstrating how coding agents transfer software engineering skills to text processing tasks.

\section{Detailed Agent Trajectories}
\label{appendix:detailed_traces}

This section provides detailed natural-language descriptions of the agent trajectories summarized in \autoref{fig:emergent_strategies}. These traces illustrate the three emergent processing strategies: iterative query refinement, programmatic aggregation, and hybrid search-read strategies.

\subsection{BrowseComp-Plus: Iterative Query Refinement}

\begin{promptbox}[Task: Multi-Hop Entity Chaining]{codingagentcolor}
Find a professional gamer satisfying: (1) American, born 2000--2010; (2) first major achievement between 2015--2023 in a game released in 2020; (3) game developed by American company founded 2000--2010; (4) one of the company's founders is married to a parenting and public health expert who earned her doctorate from a university founded in the 1930s.
\end{promptbox}

\begin{promptbox}[Agent Trajectory]{codingagentcolor}
\small
\textbf{Step 1: Initial search for question keywords.}\\
\texttt{rg "parenting and public"} across the 100K-document corpus. No direct matches found, prompting the agent to decompose the query.\\[4pt]

\textbf{Step 2: Pivot to game developer identification.}\\
\texttt{rg "Riot Games|Valorant|Marc Merrill|Brandon Beck"}\\
$\rightarrow$ Discovers Brandon Beck and Marc Merrill are co-founders of Riot Games.\\[4pt]

\textbf{Step 3: Search for founder's spouse.}\\
\texttt{rg "Brandon Beck.*married|wife"}\\
$\rightarrow$ Finds file 81813.txt mentioning Brandon Beck's marriage to Natasha Beck.\\[4pt]

\textbf{Step 4: Read file to confirm spouse details.}\\
\texttt{sed -n '1,180p' 81813.txt}\\
$\rightarrow$ Confirms Natasha Beck is ``Dr.\ Organic Mommy,'' a parenting and public health expert.\\[4pt]

\textbf{Step 5: Verify doctorate university.}\\
\texttt{sed -n '1,200p' 63541.txt}\\
$\rightarrow$ Reveals Natasha Beck holds a Psy.D.\ from Pepperdine University, founded in 1937.\\[4pt]

\textbf{Step 6: Search for the professional gamer.}\\
\texttt{rg "Demon1|Mazanov"}\\
$\rightarrow$ Finds files about Max ``Demon1'' Mazanov, American Valorant player born September 7, 2002.\\[4pt]

\textbf{Step 7: Final verification.}\\
\texttt{sed -n '1,140p' 23043.txt}\\
$\rightarrow$ Confirms Max Mazanov won Valorant Champions 2023 MVP. Valorant released 2020 by Riot Games.\\[6pt]

\textbf{Final Answer:} Max Mazanov\\[4pt]
\textbf{Key Insight:} 15+ search operations, progressively chaining entities: Riot Games $\rightarrow$ Brandon Beck $\rightarrow$ Natasha Beck $\rightarrow$ Pepperdine $\rightarrow$ Valorant $\rightarrow$ Demon1 $\rightarrow$ Max Mazanov---solving a 6-hop reasoning problem without explicit instruction.
\end{promptbox}

\subsection{Oolong-Synthetic: Programmatic Aggregation}

\begin{promptbox}[Task: NLI Classification and Aggregation]{codingagentcolor}
In a dataset of 1,772 sentence pairs, identify which user has the most instances labeled as ``contradiction.'' Labels are not provided; the agent must infer them.
\end{promptbox}

\begin{promptbox}[Agent Trajectory]{codingagentcolor}
\small
\textbf{Step 1: Initial scan to understand data structure.}\\
\texttt{sed -n '1,250p' file.txt}\\
$\rightarrow$ Discovers format: ``Date: X || User: Y || Instance: sentence1 $\rightarrow$ sentence2''. Header explains labels must be classified as entailment, neutral, or contradiction.\\[4pt]

\textbf{Step 2: Parse records with Python.}\\
Writes Python script to extract all 1,772 records, parsing user IDs and sentence pairs from the structured format.\\[4pt]

\textbf{Step 3: Build rule-based NLI classifier.}\\
Implements \texttt{contradict()} function using linguistic patterns:\\
\hspace*{4mm}$\bullet$ Negation detection: ``no'', ``not'', ``never'', ``cannot'', ``doesn't''\\
\hspace*{4mm}$\bullet$ Quantity mismatch: ``only one'' vs.\ ``series of'', ``single'' vs.\ ``many''\\
\hspace*{4mm}$\bullet$ Numerical disagreement: different numbers modifying the same noun\\
\hspace*{4mm}$\bullet$ Antonym pairs: ``can'' vs.\ ``cannot''\\[4pt]

\textbf{Step 4: Execute classifier and aggregate results.}\\
Runs classifier over all 1,772 pairs, counting contradictions per user using a \texttt{Counter} object.\\[4pt]

\textbf{Step 5: Attempt ML-based refinement.}\\
Tries to install PyTorch and Transformers for RoBERTa-MNLI classification. Installation fails due to environment conflicts; falls back to rule-based results.\\[4pt]

\textbf{Step 6: Report final answer.}\\
User 23934 identified as having the most contradiction instances.\\[6pt]

\textbf{Final Answer:} User 23934\\[4pt]
\textbf{Key Insight:} Zero search commands, $\sim$300 lines of Python code including custom NLI classifier with regex patterns. Complete abandonment of retrieval in favor of programmatic processing.
\end{promptbox}

\subsection{LongBench: Hybrid Search-Read Strategy}

\begin{promptbox}[Task: Scientific Reading Comprehension]{codingagentcolor}
What is the role of ``glacier mouse'' rolling in the warm season?\\
Choices: (A) Discharge water, (B) Get nutrients, (C) Hide Away From The Sun, (D) Preserve body heat.
\end{promptbox}

\begin{promptbox}[Agent Trajectory]{codingagentcolor}
\small
\textbf{Step 1: Initial read to understand document structure.}\\
\texttt{sed -n '1,250p'} $\rightarrow$ Identifies scientific paper about glacier mice (moss balls) on Icelandic glaciers.\\[4pt]

\textbf{Step 2: Search for keywords from question.}\\
\texttt{rg "roll|rolling|warm"} $\rightarrow$ Returns lines 359, 367, 372--390 discussing motion types: ``stick,'' ``creep,'' and ``roll.''\\[4pt]

\textbf{Step 3: Read section about rolling mechanics.}\\
\texttt{sed -n '620,700p'} $\rightarrow$ Reveals thermal properties: glacier mice provide ``thermally ameliorated environment.''\\[4pt]

\textbf{Step 4: Search for warm season context.}\\
\texttt{rg "warm season|warm"} $\rightarrow$ Finds lines about internal temperature being warmer than glacier surface.\\[4pt]

\textbf{Step 5: Search for functional role keywords.}\\
\texttt{rg "role of.*roll|function.*roll"} $\rightarrow$ Limited matches, prompting broader exploration.\\[4pt]

\textbf{Step 6: Read moss cushion dynamics section.}\\
\texttt{sed -n '930,1010p'} $\rightarrow$ Discussion of ``nutrient base necessary for plant life.''\\[4pt]

\textbf{Step 7: Search for nutrient-related content.}\\
\texttt{rg "nutrient|food|algae"} $\rightarrow$ Line 986: ``position will provide nutrient supply during the summer.''\\[4pt]

\textbf{Step 8: Read critical nutrient passage.}\\
\texttt{sed -n '600,636p'} $\rightarrow$ Confirms habitat is ``moist, relatively warm and with a ready food source.''\\[4pt]

\textbf{Step 9: Final verification.}\\
\texttt{rg "sun|shade|solar"} $\rightarrow$ Thermal protection is secondary to nutrient redistribution.\\[4pt]

\textbf{Step 10: Synthesize evidence.}\\
Concludes rolling redistributes organic matter and nutrients across the glacier surface.\\[6pt]

\textbf{Final Answer:} (B) Get nutrients\\[4pt]
\textbf{Key Insight:} 17 operations alternating between targeted searches (\texttt{rg}) and focused reading (\texttt{sed}), demonstrating hybrid strategy. Answering required synthesizing information scattered across multiple sections.
\end{promptbox}

\end{document}